\documentclass{article}

\usepackage{microtype}
\usepackage{graphicx}
\usepackage{subfigure}
\usepackage{booktabs} 

\usepackage{hyperref}

\usepackage[accepted]{icml2021}

\usepackage{lipsum}
\usepackage{adjustbox}
\usepackage{float}
\usepackage{booktabs} 
\usepackage{algorithm,algorithmic}
\usepackage{amsmath,amssymb,amsthm,bbm,mathtools}
\usepackage{multirow}
\usepackage{multicol}
\usepackage{adjustbox}

\newtheorem{theorem}{Theorem}

\DeclareMathOperator{\lrelu}{LeakyReLU}
\DeclareMathOperator{\sigmoid}{sigmoid}

\DeclareMathOperator*{\argmax}{argmax}

\newcommand{\citeseer}{\textsc{CiteSeer}}
\newcommand{\cora}{\textsc{Cora-ML}}
\newcommand{\gene}{\textsc{Gene}}

\begin{document}

\twocolumn[
\icmltitle{Generating a Doppelganger Graph: Resembling but Distinct}



\icmlsetsymbol{equal}{*}

\begin{icmlauthorlist}
\icmlauthor{Yuliang Ji}{to}
\icmlauthor{Ru Huang}{to}
\icmlauthor{Jie Chen}{goo}
\icmlauthor{Yuanzhe Xi}{to}
\end{icmlauthorlist}

\icmlaffiliation{to}{Emory University}
\icmlaffiliation{goo}{MIT-IBM Watson AI Lab, IBM Research}

\icmlcorrespondingauthor{Jie Chen}{chenjie@us.ibm.com}

\icmlkeywords{Graph generative model, deep learning}

\vskip 0.3in
]



\printAffiliationsAndNotice{}  

\begin{abstract}
Deep generative models, since their inception, have become increasingly more capable of generating novel and perceptually realistic signals (e.g., images and sound waves). With the emergence of deep models for graph structured data, natural interests seek extensions of these generative models for graphs. Successful extensions were seen recently in the case of learning from a collection of graphs (e.g., protein data banks), but the learning from a single graph has been largely under explored. The latter case, however, is important in practice. For example, graphs in financial and healthcare systems contain so much confidential information that their public accessibility is nearly impossible, but open science in these fields can only advance when similar data are available for benchmarking.

In this work, we propose an approach to generating a doppelganger graph that resembles a given one in many graph properties but nonetheless can hardly be used to reverse engineer the original one, in the sense of a near zero edge overlap. The approach is an orchestration of graph representation learning, generative adversarial networks, and graph realization algorithms. Through comparison with several graph generative models (either parameterized by neural networks or not), we demonstrate that our result barely reproduces the given graph but closely matches its properties. We further show that downstream tasks, such as node classification, on the generated graphs reach similar performance to the use of the original ones.
\end{abstract}

\section{Introduction}

\begin{table*}[t]
  \small
  \centering
  \caption{Properties of ER and BA graphs, estimated over 100 trials. ER: $n=2000$, $p=0.004$. BA: $n=2000$, $m=4$.}
  \label{graph_generator_metric_1}
  \vskip.05in
  \begin{tabular}{cc|cccccc}
    \toprule
    Model & Edge overlap
    & Clustering coef. & Charcs. path length & Triangle count & Square count \\
    \midrule
    ER & $4.03(\pm0.73)\cdot10^{-3}$ & $1.53(\pm0.16)\cdot10^{-3}$ & $3.89(\pm0.02)\cdot10^{+0}$ & $8.64(\pm0.94)\cdot10^{+1}$ & $0.00(\pm0.00)\cdot10^{+0}$ \\
    BA & $1.81(\pm0.14)\cdot10^{-2}$ & $7.38(\pm0.73)\cdot10^{-4}$ & $3.41(\pm0.02)\cdot10^{+0}$ & $6.89(\pm0.53)\cdot10^{+2}$ & $3.44(\pm1.18)\cdot10^{+1}$ \\
    \bottomrule
  \end{tabular}
  \begin{tabular}{c|ccccccc}
    \toprule
    Model & LCC & Powerlaw exponent & Wedge count & Rel. edge distr. entr. & Gini coefficient \\
    \midrule
    ER & $2.00(\pm0.00)\cdot10^{+3}$ & $1.50(\pm0.00)\cdot10^{+0}$ & $6.38(\pm0.14)\cdot10^{+4}$ & $9.92(\pm0.00)\cdot10^{-1}$ & $1.96(\pm0.04)\cdot10^{-1}$ \\
    BA & $2.00(\pm0.00)\cdot10^{+3}$ & $3.13(\pm0.12)\cdot10^{+0}$ & $1.45(\pm0.06)\cdot10^{+5}$ & $9.58(\pm0.01)\cdot10^{-1}$ & $3.64(\pm0.02)\cdot10^{-1}$ \\
    \bottomrule
  \end{tabular}
\end{table*}

The advent of deep learning stimulated rapid progress of deep generative models. These models, parameterized by neural networks, are capable of learning the complex distribution of a set of training data and generating new samples from it. Such deep generative models have been successfully applied in various applications, including paintings~\citep{Oord2016}, sound tracks~\citep{Oord2016a}, and natural languages~\citep{Hu2017}. The success, in part, is owing to the flourishing development of neural network architectures that effectively decode Euclidean-like data such as images and sequences, notable architectures including CNNs, RNNs, attention mechanisms, and transformers.

Graphs are a special, yet extensively used, data type. They model pairwise relationships among entities. Their realizations range from social networks, transaction networks, power grids, to program structures and molecules. Generative models for graphs emerged as research on deep learning for graph structured data gained momentum recently. They were facilitated by a plethora of graph neural networks~\citep{Li2016,Gilmer2017,Kipf2017,Hamilton2017,Velickovic2018,Xu2019} that effectively learn graph representations and sophisticated decoding architectures~\citep{Simonovsky2018,Grover2019,Johnson2017,graphrnn2018you,gran2019liao} that map continuous vectors to combinatorial graph structures.

The majority of deep generative models for graphs~\citep{Jin2018,Ma2018,Simonovsky2018,Grover2019,graphrnn2018you,gran2019liao} follows the paradigm of ``learning from a distribution of graphs and generating new ones.'' Another scenario, which bears equal significance in practice, is to learn from a single graph and generate new ones. This scenario is largely under explored. The scenario meets graphs that are generally large and unique, which contains rich structural information but rarely admits replicas. Examples of such graphs are financial networks, power networks, and domain-specific knowledge graphs. Often, these graphs come from proprietary domains that prevent their disclosure to the public for open science. Hence, machine learning methods are desired to produce non-harmful surrogates that carry the essential graph properties for model benchmarking but otherwise disclose no confidential information and prohibit reverse engineering of the original graph.

One live example that necessitates the creation of surrogate graphs is in the financial industry. A financial entity (e.g., banks) maintains a sea of transaction data, from which traces of fraud are identified. Graph neural network techniques have been gaining momentum for such investigations~\citep{Weber2018,Weber2019}, wherein a crucial challenge is the lack of realistic transaction graphs that no existing graphs resemble and that synthetic generators fail to replicate. On the other hand, many barriers exist for the financial entities to distribute their data, even to its own departments or business partners, because of levels of security and privacy concerns. Our work aims to generate high quality surrogates, for which the research community can develop suitable models, which in turn financial entities are confident to use.

We propose an approach to learning the structure of a given graph and constructing new ones that carry a similar structure. The new graphs must not reproduce the original one, since otherwise the learning becomes a memorization and it is less useful in practice. We call the generated graph \emph{doppelganger}, which (i) matches the input graph in many characteristics but (ii) can hardly be used to infer the input. Both aims can be quantified. For (i), we will measure a large number of graph properties; they need be close. For (ii), we will measure the edge overlap between two graphs; it should be nearly zero.

It is worthwhile to distinguish our setting from that of NetGAN~\citep{Bojchevski18NetGAN}, a recent and representative approach to learning from a single graph. NetGAN solves the problem by learning the distribution of random walks on the given graph; then, it assembles a graph from new random walks sampled from the learned model. With more and more training examples, the generated graph is closer and closer to the original one with increasing edge overlap, which goes in the opposite direction of aim (ii) above. In essence, NetGAN learns a model that memorizes the transition structure of the input graph and does not aim to produce a separate graph. This approach capitalizes on a fixed set of nodes and determines only edges among them, as most of the prior work under the same setting does. We intend to generate new nodes on the contrary.

A natural question asks if it is possible to produce a different graph that barely overlaps with a given one but matches its properties. The answer is affirmative and easy examples may be found from random graphs. Intuitively, a random graph model describes a class of graphs that share similar properties; and because of randomness, it is possible that one does not overlap with the other at all. For example, an Erd\"{o}s--R\'{e}nyi (ER) graph model~\citep{Erdoes1959,Gilbert1959} with node count $n$ and edge probability $p$ produces two graphs with an expected edge overlap $p$. The overlap can be made arbitrarily small when $p\to0$. On the other hand, their graph properties can be quite close. In Table~\ref{graph_generator_metric_1}, we list, together for the Barab\'{a}si--Albert (BA) random graph model~\citep{Barabasi1999}, estimates of several graph properties. For each statistic, the standard deviation is generally $10\%$ (and sometimes $1\%$ or even smaller) of the mean, indicating that the property value of two random graphs does not differ much. Meanwhile, the ER graphs have an edge overlap approximately $0.004$ and the BA graphs $0.018$. These values can be arbitrarily small when one adjusts the model parameters.

Hence, we proceed to develop a method that aims at producing new graphs that match the properties of an input graph but minimally overlap with it. The method leverages several deep learning ingredients, including graph representation learning and generative adversarial networks, as well as a graph realization algorithm from combinatorics. We perform experiments on real-life data sets to demonstrate that the generated graphs match the properties and produce similar node classification results to the original graphs.

\section{Related Work}
Early generative models for graphs focus on characterizing certain properties of interest. The Erd\"{o}s--R\'{e}nyi model~\citep{Erdoes1959,Gilbert1959}, dating back to the 1950s, is one of the most extensively studied random models. The Watts--Strogatz model~\citep{Watts1998} intends to capture the small-world phenomenon seen in social networks; the Barab\'{a}si--Albert model~\citep{Barabasi1999} mimics the scale-free (powerlaw) property; and the stochastic block model~\citep{Goldenberg2010,dcsbm2011} models community structures. These models, though entail rich theoretical results, are insufficient to capture all aspects of real-life graphs.

The complex landscape of graphs in applications urges the use of higher-capacity models, predominantly neural networks, to capture all properties beyond what mathematical modeling covers. Two most widely used frameworks are VAEs~\citep{Kingma2014} and GANs~\citep{goodfellow2014gan}. The VAE framework encodes a graph into a latent representation and decodes it for maximal reconstruction. Approaches such as GraphVAE~\citep{Simonovsky2018} and Graphite~\citep{Grover2019} reconstruct the graph adjacency matrix (and labels if any). The Junction Tree VAE approach~\citep{Jin2018} leverages knowledge of molecular substructures and uses the tree skeleton rather than the original graph for encoding and decoding. To handle general structural constraints, \citet{Ma2018} formulate them as regularizations in the variational lower bound objective.

The GAN framework uses a generator to produce graphs (typically in the form of adjacency matrices) and a discriminator to tell produced results from training examples. The MolGAN approach~\citep{DeCao2018} incorporates a reinforcement learning term into the discriminator loss to promote molecule validity. The LGGAN approach~\citep{lggan2019fan} additionally generates graph-level labels. Besides decoding a graph adjacency matrix, one may treat the graph generation process as sequential decision making, producing nodes and edges one by one. Representative examples of RNN-type of graph decoders are DeepGMG~\citep{Li2018}, GraphRNN~\citep{graphrnn2018you}, and GRAN~\citep{gran2019liao}.

All the above methods learn from a collection of graphs. More relevant to this paper is the scenario that the training set consists of one single graph. NetGAN~\citep{Bojchevski18NetGAN} is one of the rare studies under this scenario. This approach reformulates the problem as learning from a collection of random walks, so that a new graph is generated through assembling newly sampled random walks. VGAE~\citep{kipf2016variational} also learns from a single graph; but its aim is to reconstruct the graph rather than producing new ones.

\begin{figure*}[ht]
  \centering
  \includegraphics[width=0.9\linewidth]{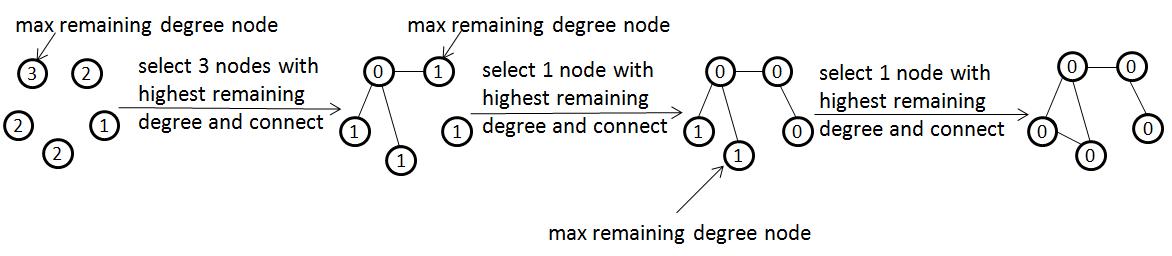}
  \includegraphics[width=0.9\linewidth]{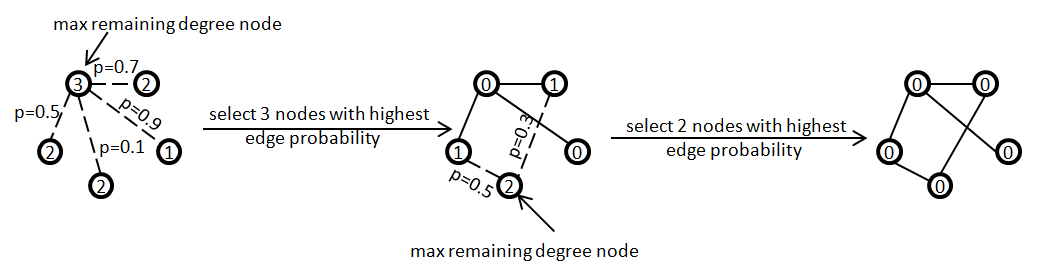}
  \caption{Examples of generating a graph by using the original HH (top) and the proposed improved HH algorithm (bottom).}
  \label{HH_algorithm_process}
\end{figure*}

\section{Method}
In this section, we describe the proposed approach to generating a doppelganger graph $G$ given an input graph $G_0$, by leveraging the Havel--Hakimi (HH) algorithm. We  improve HH through using GAN to generate nodes distinct from those of $G_0$ and leveraging a link predictor to guide the neighbor selection process of HH.

\subsection{Graph Realization Algorithm: Havel--Hakimi}
A sequence $S$ of integers $(d_1,d_2,\ldots,d_n)$ is called \emph{graphic} if the $d_i$'s are nonnegative and they are the degrees of a graph with $n$ nodes. Havel--Hakimi~\citep{Havel1955,Hakimi1962} is an algorithm that determines whether a given sequence is graphic. The algorithm is constructive in that it will construct a graph whose degrees coincide with $S$, if graphic.

At the heart of HH is the following theorem: Let $S$ be nonincreasing. Then, $S$ is graphic if and only if $S'=(d_2-1,d_3-1,\ldots,d_{d_1+1}-1,d_{d_1+2},\ldots,d_n)$ is graphic. Note that $S'$ is not necessarily sorted. This theorem gives a procedure to attempt to construct a graph whose degree sequence coincides with $S$: Label each node with the desired degree. Beginning with the node $v$ of highest degree $d_1$, connect it with $d_1$ nodes of the next highest degrees. Set the remaining degree of $v$ to be zero and reduce those of the $v$'s neighbors identified in this round by one. Then, repeat: select the node with the highest remaining degree and connect it to new neighbors of the next highest remaining degrees. The procedure terminates when (i) one of the nodes to be connected to has a remaining degree zero or (ii) the remaining degrees of all nodes are zero. If (i) happens, then $S$ is not graphic. If (ii) happens, then $S$ is graphic and the degree sequence of the constructed graph is $S$. The pseudocode is given in Algorithm~\ref{algorithm_Havel-Hakimi}. The top panel of Figure~\ref{HH_algorithm_process} illustrates an example.

\begin{algorithm}[ht]
  \caption{Havel--Hakimi Algorithm}
  \label{algorithm_Havel-Hakimi}
  \begin{algorithmic}
    \STATE {\bfseries Input:} Graphic sequence $\{d_i\}$
    \STATE Initialize $adj\_list(i)=\{\}$ for all nodes $i$
    \REPEAT
    \STATE Initialize $rd_i=d_i-len(adj\_list(i))$ for all $i$
    \STATE Set $done=true$
    \STATE Select node $k=\argmax_i(rd_i)$
    \REPEAT
    \STATE Select node $t=\argmax_{i\ne k, \,\, i\notin adj\_list(k)}(rd_i)$
    \IF{$rd_t>0$}
    \STATE Add $t$ to $adj\_list(k)$ and add $k$ to $adj\_list(t)$
    \STATE Set $done=false$
    \ENDIF
    \UNTIL{$rd_t=0$ or $d_k=len(adj\_list(k))$}
    \UNTIL{$done$ is $true$ or $len(adj\_list(i))=d_i$ for all $i$}
  \end{algorithmic}
\end{algorithm}

The most attractive feature of HH is that the constructed graph reproduces the degree sequence of the original graph, thus sharing the same degree-based properties, as the following statement formalizes.

\begin{theorem}\label{theorem 1}
Let $P(\{d_1,\ldots,d_n\})$ be a graph property that depends on only the node degrees $d_1$, \ldots, $d_n$. A graph $G$ realized by using the Havel--Hakimi algorithm based on the degree sequence of $G_0$ has the same value of $P$ as does $G_0$. Examples of $P$ include wedge count, powerlaw exponent, entropy of the degree distribution, and the Gini coefficient.
\end{theorem}

On the other hand, a notable consequence of HH is that the constructed graph tends to generate large cliques.

\begin{theorem}\label{theorem 2}
Suppose the degree sequence of a given graph is $d_1 \geq d_2 \geq ... \geq d_n$. If there exists an integer $k$ such that $d_k-d_{k+1} \geq k-1$, then the graph realized by the Havel--Hakimi algorithm has a clique with at least $k$ nodes.
\end{theorem}

Theorem~\ref{theorem 2} indicates a potential drawback of HH. When connecting with neighbors, HH always prefers high-degree nodes. Thus, iteratively, these nodes form a large clique (complete subgraph), which many real-life graphs lack. Furthermore, since any induced subgraph (say, with $m$ nodes) of a $k$-clique is also a clique, the number of $m$-cliques grows exponentially with $k$. Then, the number of small motifs (e.g., triangles and squares) easily explodes, departing substantially from the characteristics of a real-life graph. In other words, graphs generated by HH possibly fail to preserve certain local structures of the original graph.


\subsection{Improved Havel--Hakimi}
To overcome the drawback of HH, we propose two improvements. First, rather than selecting neighbors in each round by following a nonincreasing order of the remaining degrees, we select neighbors according to link probabilities. Second, such probabilities are computed by using a link prediction model together with new nodes sampled from the node distribution of the original graph $G_0$. This way, the new graph $G$ carries the node distribution and the degree sequence information of $G_0$ but has a completely different node set, leading to good resemblance but low overlap.

In what follows, we elaborate the two improvements in order. We first describe the change to HH. In each iteration, a node $k$ with maximal remaining degree is first selected. This node is then connected to a node $t$ with the highest link probability $p_{kt}$ among all nodes $t$ not being connected but with a nonzero remaining degree. Connect as many such $t$ as possible to fill the degree requirement of $k$. If $k$ cannot find enough neighbors (no more $t$ exists with nonzero remaining degree),  the algorithm will skip node $k$. This process is repeated until no more nodes can add new neighbors. The pseudocode is given in Algorithm~\ref{algorithm_twinGraph}. The bottom panel of Figure~\ref{HH_algorithm_process} illustrates an example.

\begin{algorithm}[ht]
  \caption{Improved Havel--Hakimi (proposed method)}
  \label{algorithm_twinGraph}
  \begin{algorithmic}
    \STATE {\bfseries Input:} Graphic sequence $\{d_i\}$ 
    \STATE Initialize $adj\_list(i)=\{\}$ for all nodes $i$
    \REPEAT
    \STATE Initialize $rd_i=d_i-len(adj\_list(i))$ for all $i$
    \STATE Set $done=true$
    \STATE Select node $k=\argmax_i(rd_i)$
    \REPEAT
    \STATE Select node $t=\argmax_{i\notin adj\_list(k), \,\, rd_i>0}(p_{ki})$,
    \STATE \quad where $p_{ki}=link\_prediction(k,i)$
    \STATE If no such $t$ exists, exit loop
    \STATE Add $t$ to $adj\_list(k)$ and add $k$ to $adj\_list(t)$
    \STATE Set $done=false$
    \UNTIL{$len(adj\_list(k))=d_k$}
    \UNTIL{$done$ is $true$ or $len(adj\_list(i))=d_i$ for all $i$}
  \end{algorithmic}
\end{algorithm}

Next, we explain how the probabilities for neighbor selection are computed.

\begin{figure*}[t]
  \centering
  \includegraphics[width=.8\linewidth]{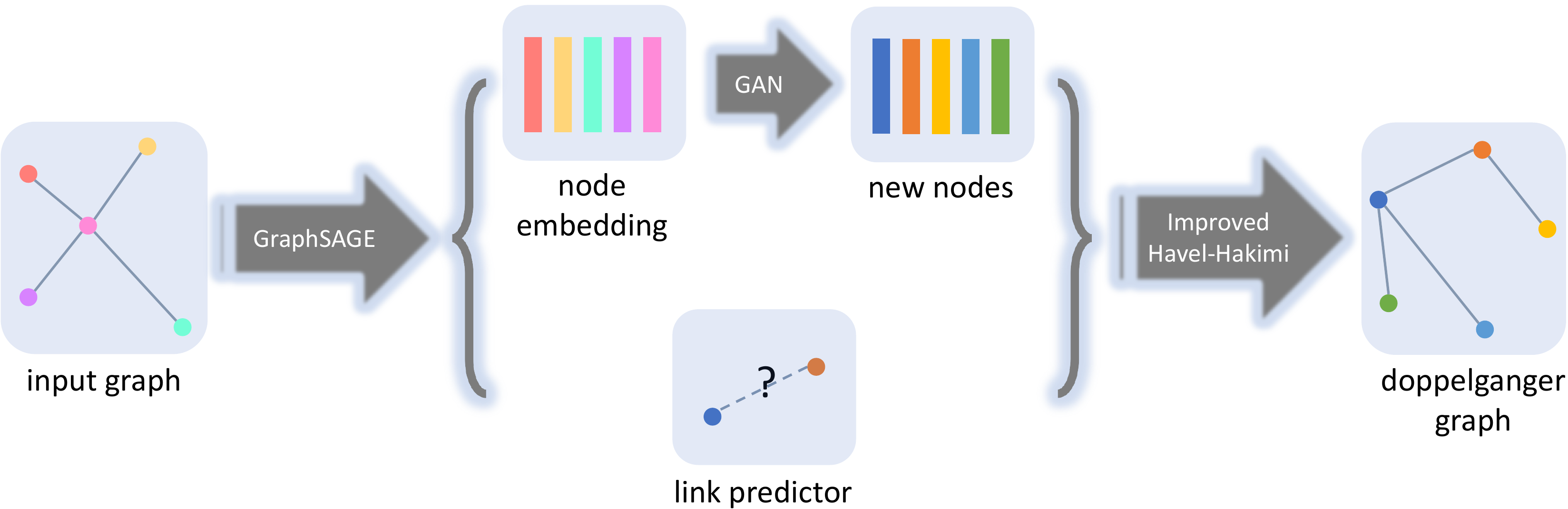}
  \caption{Pipeline to generate a doppelganger graph.}
  \label{TwinGraph_architecture}
\end{figure*}

\textbf{Link prediction model.}
Probabilities naturally come from a link prediction model, which dictates how likely an edge is formed. We opt to use a graph neural network (GNN) trained with a link prediction objective to obtain the link predictor. The GNN produces, as a byproduct, node embeddings that will be used subsequently for sampling new nodes. There exist many non-neural network approaches~\citep{Perozzi2014,Tang2015,grover2016node2vec} for producing node embeddings, but they are not trained with a link predictor simultaneously.

We use GraphSAGE~\citep{Hamilton2017}, which offers satisfactory link prediction performance, but enhance the standard link prediction model $p=\sigmoid(z_{v}^Tz_{u})$ for two nodes $v$ and $u$ by a more complex one:
\[
p=\sigmoid(W_{2}\cdot\lrelu(W_{1}(z_v \circ z_u)+b_1)+b_2),
\]
where $W_1,b_1,W_2,b_2$ are parameters trained together with GraphSAGE. This link prediction model is slightly parameterized and we find that it works better for larger graphs.

\textbf{Sampling new nodes.}
After GraphSAGE is trained, we obtain as byproduct node embeddings of the input graph. These embeddings form a distribution, from which one may sample new embeddings. They form the node set of the doppelganger graph. Sampling may be straightforwardly achieved by using a deep generative model; we opt to use Wasserstein GAN~\citep{Arjovsky2017}.
We train WGAN by using the gradient penalty approach proposed by~\citet{Ishaan2017ImprovedGan}.

\textbf{Node labeling with degree sequence.}
A remaining detail is the labeling of the new nodes with the desired degree. We apply the link predictor on the new nodes to form an initial doppelganger graph. With this graph, nodes are ordered by their degrees. We then label the ordered nodes by using the degree sequence of the input graph. This way, the improved HH algorithm will run with the correct degree sequence and result in the final doppelganger graph.

\subsection{Summary}
To summarize, given an input graph $G_0$, our method first trains a GraphSAGE model on $G_0$ with a link prediction objective, yielding node embeddings and a link predictor. We then train a Wasserstein GAN on the node embeddings of $G_0$ and sample new embeddings from it. Afterward, we run the improved HH algorithm, wherein needed link probabilities are calculated by using the link predictor on the new embeddings. The resulting graph is a doppelganger of $G_0$. See Figure~\ref{TwinGraph_architecture}.

\section{Experiments}
In this section, we perform comprehensive experiments to show that the proposed method generates graphs closely matching the input ones in a wide range of properties yet barely overlapping with them. We also perform a downstream task---node classification---on the generated graphs and demonstrate that they can be used as surrogates of the input graphs with similar classification performance.

\subsection{Setup}

\textbf{Data sets.}
We experiment with three commonly used data sets for graph based learning:
\cora~\citep{McCallum2000}\footnote{\url{https://github.com/danielzuegner/netgan}},
\citeseer~\citep{Sen2008}\footnote{\url{https://linqs.soe.ucsc.edu/data}},
and~\gene~\citep{Rossi2015}\footnote{\url{http://networkrepository.com/gene.php}}.
As is the usual practice, we experiment with the largest connected component of each graph. Their information is summarized in Table~\ref{tab:dataset}.

\begin{table}[ht]
  \centering
  \small
  \caption{Data sets (largest connected component).}
  \label{tab:dataset}
  \vskip.05in
  \begin{tabular}{lcccc}
    \toprule
    Name & \# Nodes & \# Edges & \# Classes & \# Features\\
    \midrule
    \cora     & 2810 & 6783 & 7 & 1433\\
    \citeseer & 2120 & 3679 & 6 & 3703\\
    \gene     & 814  & 1436 & 2 & N/A\\
    \bottomrule
  \end{tabular}
\end{table}

\textbf{Graph generative models.}
We compare with a number of graph generative models, either deep or non-deep learning based. For non-deep learning based models, we consider 
the configuration model (CONF)~\citep{Molloy1995},
the Chung-Lu model~\citep{ChungLu2001},
the degree-correlated stochastic block model (DC-SBM)~\citep{dcsbm2011}, and
the exponential random graph model (ERGM)~\citep{ERGM1981}.
For deep learning based models, we consider
variational graph autoencoder (VGAE)~\citep{kipf2016variational} and
NetGAN~\citep{Bojchevski18NetGAN}.
We use either off-the-shelf software packages or the codes provided by the authors to run these models. The choices of (hyper)parameters mostly follow default if they are provided, or tuned if not. For details, see Section~\ref{sec:baseline}.

It is important to note that the deep learning models VGAE and NetGAN desire a high edge overlap by the nature of their training algorithms. This goal is at odds with our purpose, which aims at little overlap for counteracting reverse engineering. Moreover, the way CONF works is that one specifies a desired edge overlap, retains a portion of edges to satisfy the overlap, and randomly rewires the rest while preserving the node degrees. The random rewiring succeeds with a certain probability that increases as overlap increases. Hence, to ensure that CONF successfully returns a graph in a reasonable time limit, the overlap cannot be too small, also at odds with our purpose.

\textbf{Graph properties.}
We use a number of graph properties to evaluate the closeness of the generated graph to the original graph. Nine of them are global properties: clustering coefficient, characteristic path length, triangle count, square count, size of the largest connected component, powerlaw exponent, wedge count, relative edge distribution entropy, and Gini coefficient. These properties are also used by~\citet{Bojchevski18NetGAN} for evaluation.

Additionally, we measure three local properties: local clustering coefficients, degree distribution, and local square clustering coefficients. They are all distributions over nodes and hence we compare the maximum mean discrepancy (MMD) of the distributions between the generated graph and the original graph. For details of all properties, see Section~\ref{sec:property}.

\textbf{Our model configuration and training.}
We use a single WGAN architecture for all graphs to learn embedding distributions. We also enhance GraphSAGE training to obtain satisfactory link prediction performance. The training procedure and the architectures are detailed in Section~\ref{sec:param}.

\subsection{Results}

\textbf{Comparison with HH.}
We first compare the original HH algorithm and our improved version. As elucidated earlier (see Theorem~\ref{theorem 2}), a potential drawback of HH is that it tends to generate large cliques, which lead to an exploding number of small motifs that mismatch what real-life graphs entail. The improved HH, on the other hand, uses link probabilities to mitigate the tendency of forming large cliques. Table~\ref{coraml-hhcompare} verifies that improved HH indeed produces triangle (3-clique) counts and square (4-clique) counts much closer to those of the original graph. One also observes the exploding square count on \cora\ and \citeseer\ by using the original HH algorithm, corroborating the tendency of large cliques in large graphs.

\begin{table}[ht]
  \small
  \centering
  \caption{Comparison of HH and improved HH. In the brackets are standard deviations (based on five random repetitions).}
  \label{coraml-hhcompare}
  \vskip.05in
  \begin{tabular}{ccc}
    \toprule
    & Triangle count & Square count\\
    \midrule 
    \cora       & $2810$ & $517$ \\
    \midrule
    HH          & $5420(90)$ & $16900(200)$ \\
    Improved HH & $3060(200)$ & $2220(560)$\\
    \bottomrule
  \end{tabular}
  \begin{tabular}{ccc}
    \toprule
    & Triangle count & Square count \\
    \midrule 
    \citeseer   & $1080$ & $249$ \\
    \midrule
    HH          & $1120(20)$ & $1810(50)$\\
    Improved HH & $700(48)$ & $210(72)$ \\
    \bottomrule
  \end{tabular}
  \begin{tabular}{ccc}
    \toprule
    & Triangle count & Square count \\
    \midrule 
    \gene       & $809$ & $968$ \\
    \midrule
    HH          & $351(6)$ & $351(7)$\\
    Improved HH & $654(19)$ & $417(73)$ \\
    \bottomrule
  \end{tabular}
\end{table}

\textbf{Quality of generated graphs.}
We now compare the graphs generated by our improved HH algorithm with those by other generative models introduced in the preceding subsection. Table~\ref{metric_table_cora} summarizes the results on \cora\ in both numeric format and chart format. Results of \citeseer\ and \gene\ follow similar observations and they are left in the supplement (Section~\ref{sec:remain}).

\begin{table*}[ht!]
  \small
  \centering
  \caption{Properties of \cora\ and the graphs generated by different models (averaged over five random repetitions). EO means edge overlap and numbers in the bracket are standard deviations.}
  \label{metric_table_cora}
  \vskip.05in
  \begin{adjustbox}{max width=\textwidth}
    \begin{tabular}{c@{\hspace{.5em}}ccccccc}
      \toprule
      & EO & Cluster. coeff.  & Charcs. path length & Triangle count & Square count & LCC & Powerlaw exponent \\
      & & $\times10^{-3}$& $\times10^{+0}$& $\times10^{+3}$& $\times10^{+2}$&$\times10^{+3}$&$\times10^{+0}$\\
      \midrule
      \cora       &          & $2.73$ & $5.61$ & $2.81$& $5.17$ & $2.81$ & $1.86$ \\
      \midrule
      CONF        & $42.4\%$ & $0.475(0.023)$ & $4.47(0.02)$ & $0.490(0.024)$ & $0.094(0.049)$ & $\boldsymbol{2.79(0.00)}$ & $\boldsymbol{1.86(0.00)}$ \\
      DC-SBM      & $4.88\%$ & $2.26(0.13)$ & $4.59(0.02)$ & $1.35(0.05)$& $\boldsymbol{0.923(0.015)}$ & $2.48(0.03)$ & $5.35(0.06)$  \\
      Chung-Lu    & $1.29\%$ & $0.548(0.0583)$ & $4.08(0.01)$ & $0.550(0.032)$ & $0.324(0.106)$ & $2.47(0.01)$ & $1.79(0.01)$\\
      ERGM        & $4.43\%$ & $1.53(0.08)$ & $4.86(0.01)$ & $0.0586(0.01)$ & $0.00(0.00)$ & $\boldsymbol{2.79(0.00)}$ & $1.65(0.00)$\\
      VGAE        & $53.1\%$ & $6.30(0.24)$ & $5.16(0.09)$ & $13.4(0.3)$ & $2320(120)$ & $1.98(0.04)$ & $1.82(0.00)$\\
      NetGAN      & $49.9\%$ & $3.75(0.94)$ & $4.30(0.30)$ & $12.5(0.5)$ & $184(19)$ & $1.96(0.07)$ & $1.77(0.01)$\\
      \midrule
      Improved HH & $\boldsymbol{0.20\%}$ & $\boldsymbol{2.97(0.20)}$ & $\boldsymbol{5.67(0.11)}$ & $\boldsymbol{3.06(0.20)}$ & $22.2(5.6)$ & $2.52(0.02)$ & $\boldsymbol{1.86(0.00)}$\\
      \bottomrule
      \toprule
      & EO & Wedge count & Rel. edge distr. entr. & Gini coefficient& Local cluster. & Degree distr. &  Local sq. cluster. \\
      &  & $\times10^{+5}$ & $\times10^{-1}$ & $\times10^{-1}$ & $\times10^{-2}$ & $\times10^{-2}$ &  $\times10^{-3}$\\
      \midrule
      \cora       &          & $1.02$ & $9.41$ & $4.82$       & 0 & 0 & 0\\
      \midrule
      CONF        & $42.4\%$ & $\boldsymbol{1.02(0.00)}$ & $\boldsymbol{9.41(0.00)}$ & $\boldsymbol{4.82(0.00)}$ & $4.76(0.04)$ & $\boldsymbol{0}$ & $3.08(0.07)$\\
      DC-SBM      & $4.88\%$ & $0.923(0.015)$ & $9.30(0.02)$ & $5.35(0.06)$ & $4.40(0.09)$ & $1.55(0.08)$ & $2.89(0.11)$\\
      Chung-Lu    & $1.29\%$ & $1.08(0.03)$ & $9.26(0.01)$ & $5.46(0.02)$ & $5.53(0.07)$ & $1.66(0.18)$ & $3.29(0.03)$\\
      ERGM        & $4.43\%$ & $0.426(0.007)$ & $9.84(0.00)$ & $2.67(0.02)$& $6.09(0.03)$ & $9.67(0.08)$ & $3.38(0.00)$ \\
      VGAE        & $53.1\%$ & $1.79(0.02)$ & $8.73(0.00)$ & $7.03(0.01)$ & $4.44(0.27)$ & $10.25(0.28)$ & $9.65(0.69)$\\
      NetGAN      & $49.9\%$ & $2.08(0.18)$ & $8.70(0.06)$ & $7.04(0.15)$& $4.01(0.63)$ & $13.13(1.77)$ & $5.52(3.18)$ \\
      \midrule
      Improved HH & $\boldsymbol{0.20\%}$ & $\boldsymbol{1.02(0.00)}$ & $\boldsymbol{9.41(0.00)}$ & $\boldsymbol{4.82(0.00)}$ & $\boldsymbol{3.91(0.10)}$ & $\boldsymbol{0}$ & $\boldsymbol{2.08(0.07)}$ \\
      \bottomrule
    \end{tabular}
  \end{adjustbox}

  \vskip30pt
  \includegraphics[width=\linewidth]{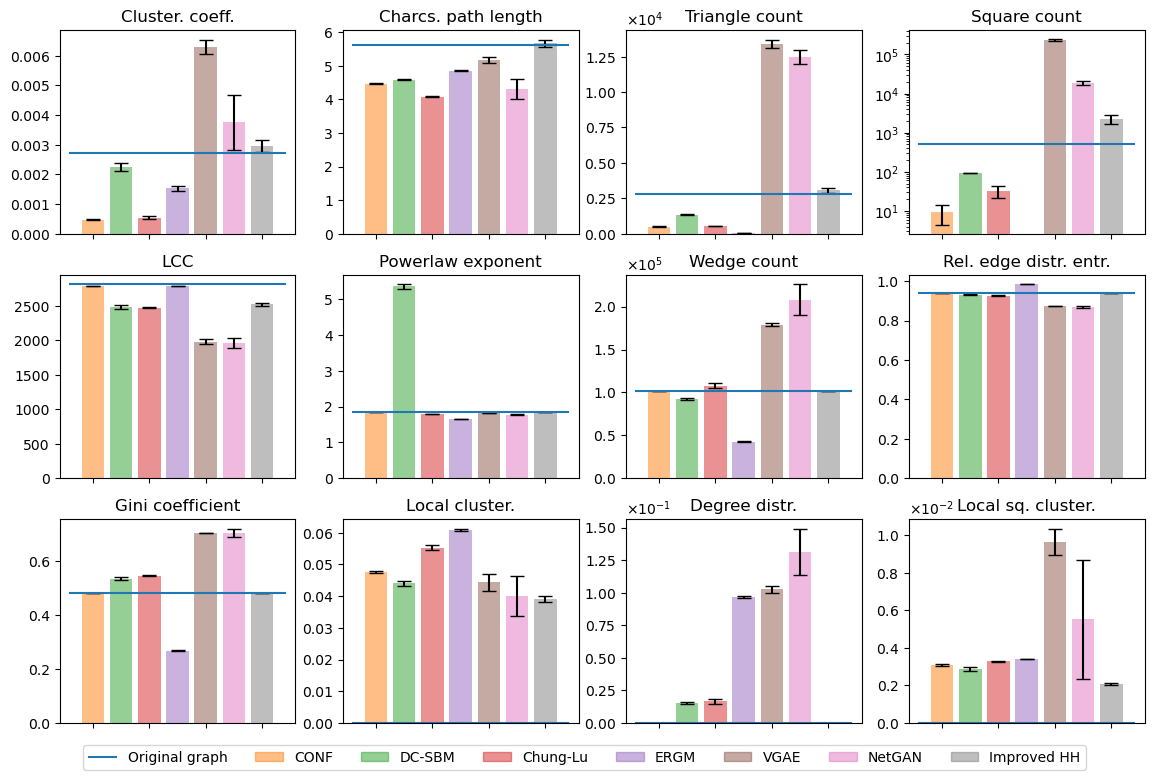}
\end{table*}

\begin{figure*}[ht]
  \centering
  \includegraphics[width=.32\linewidth]{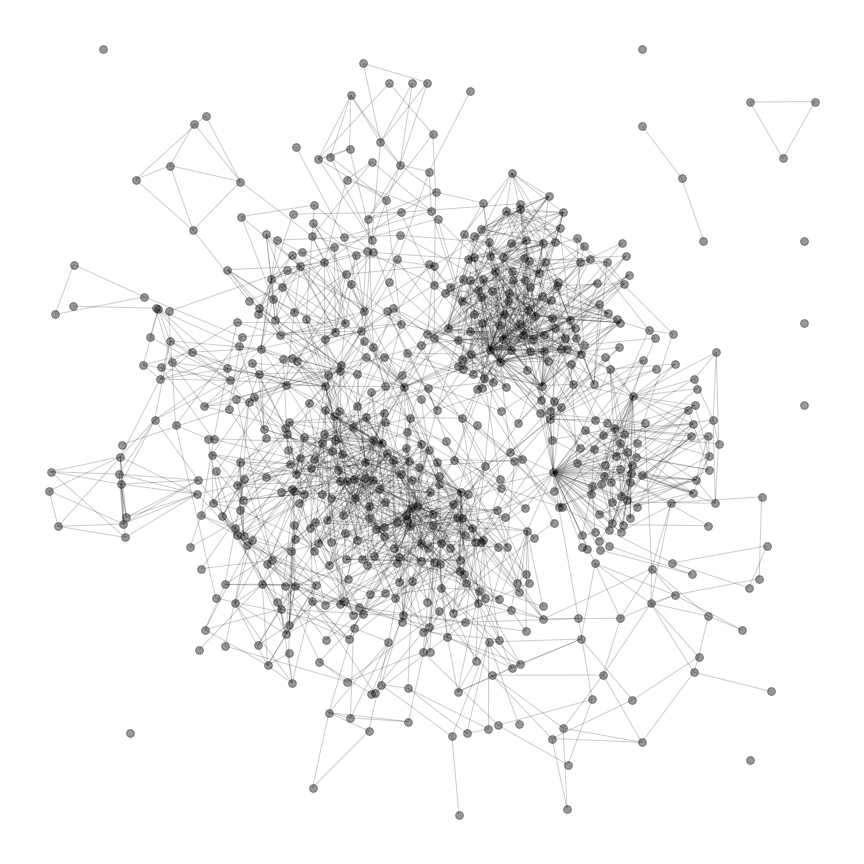}
  \includegraphics[width=.32\linewidth]{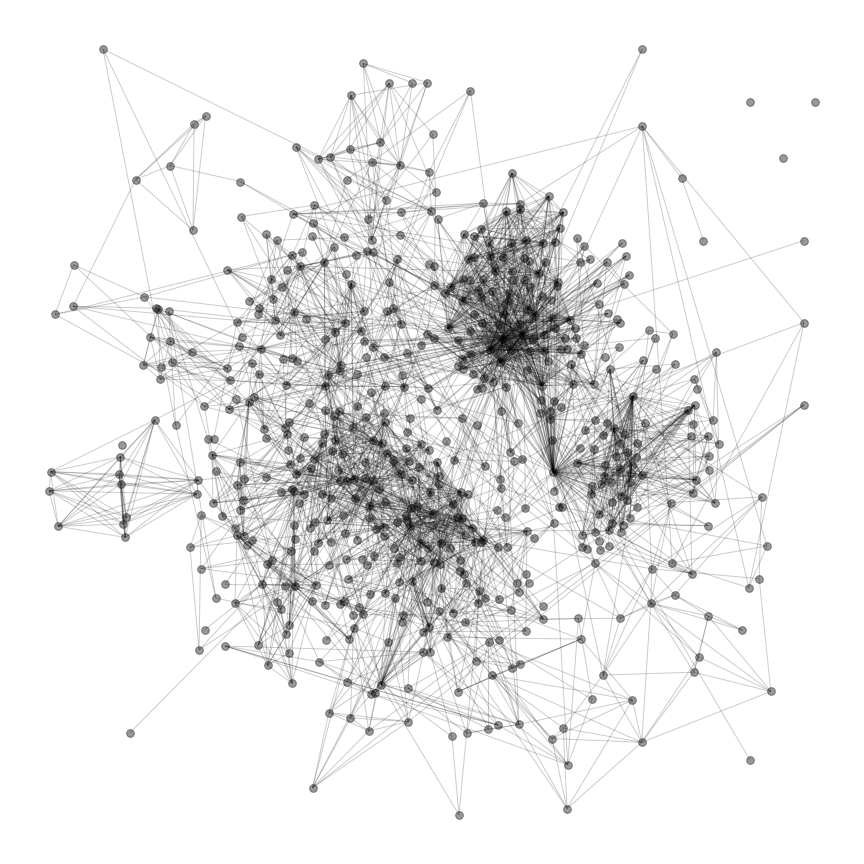}
  \includegraphics[width=.32\linewidth]{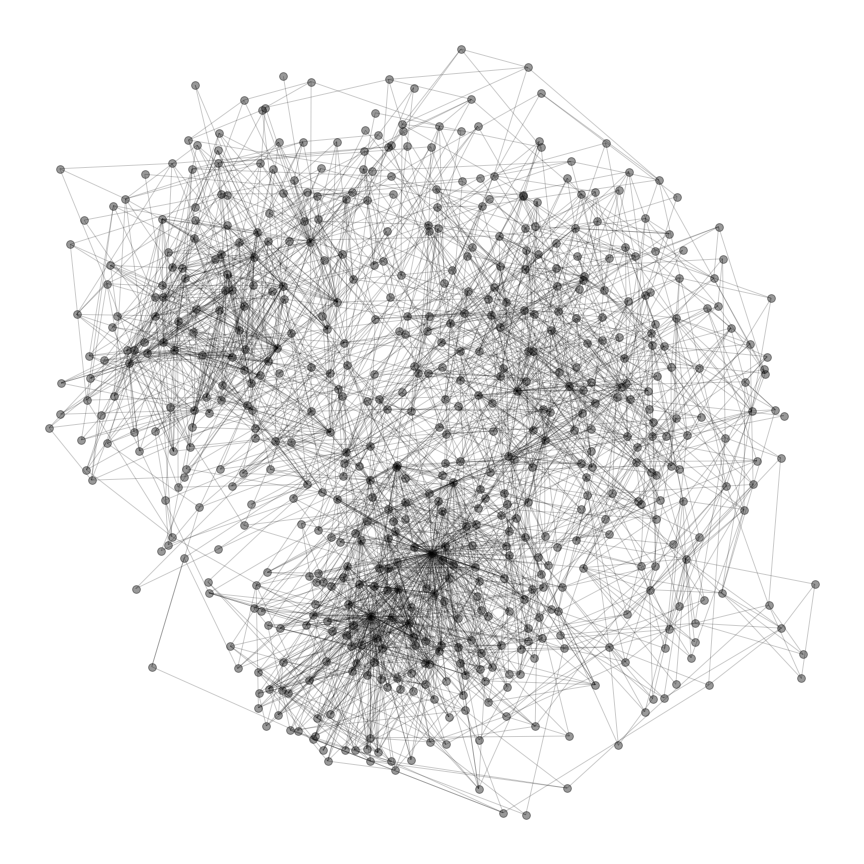}
  \caption{\cora. Left: original graph; middle: generated by NetGAN; right: generated by improved HH. Drawn are the subgraphs induced by the first 600 nodes ordered by degrees.}
  \label{cora_result}
\end{figure*}

First, our model results in the lowest edge overlap, fulfilling a goal opposite to that of VGAE and NetGAN. Even considering other models that do not aim at optimizing the overlap toward either end, our model performs desirably. The generated graphs guard against adversarial exploitation of the original edge information.

Second, for most of the global properties considered, our model produces graphs closest to the original graph, with a few properties exactly matched. The exact matching (powerlaw exponent, wedege count, relative edge distribution entropy, and Gini coefficient) is not surprising in light of Theorem~\ref{theorem 1}, because these properties are solely degree based. Furthermore, for other non-degree based properties (clustering coefficients, characteristic path length, triangle count, square count, and largest connected component), a close match indicates that our model learns other aspects of the graph structure reasonably well. Note that CONF also matches the degree-based properties, by design. However, CONF results are difficult to obtain, because the random wiring hardly successfully produces a graph matching the degree sequence when the prescribed overlap is small.

Third, for local properties (local clustering coefficients, degree distribution, and local square clustering coefficients), graphs from our model generally are closest to the original graphs, than are those from other models. For our model, the MMD of the degree distributions is zero according to Theorem~\ref{theorem 1}. On the other hand, that the MMD of the distributions of the clustering coefficients is the smallest (or close to smallest) manifests that local characteristics of the input graph are well learned.

\textbf{Downstream example: node classification.}
In practice, the generated graph is used as a surrogate of the original one in downstream tasks. For model developers, they desire that the model performs similarly on both graphs, so that both the graph owner and the developer are confident of the use of the model on the original graph. We experiment with the node classification task and test with more than one model: GCN~\citep{Kipf2017} and GraphSAGE~\citep{Hamilton2017}. Table~\ref{node_classification} shows that the graphs generated by our method indeed satisfy the desire, producing competitively similar classification accuracies; whereas those generated by other methods fall short, with much poorer accuracies. For experimentation protocol and the setup, see Section~\ref{sec:node.classi} in the supplement.

\begin{table}[ht]
  \small
  \centering
  \caption{Node classification results (10-fold cross validation) on original/generated graphs. Top: GCN; bottom: GraphSAGE.}
  \label{node_classification}
  \vskip.05in
  \begin{adjustbox}{max width=\linewidth}
    \begin{tabular}{cccc}
      \toprule
      & \cora & \citeseer & \gene \\
      \midrule
      Origin. graph & $76.37(0.07)$ & $84.11(0.07)$ & $81.46(0.12)$\\
      \midrule
      CONF        & $35.61(0.12)$ & $51.99(0.06)$ & $64.37(0.69)$\\
      DC-SBM      & $44.29(0.18)$ & $50.06(0.20)$ & $64.25(1.82)$\\
      Chung-Lu    & $35.61(0.12)$ & $36.10(0.08)$ & $58.99(1.24)$\\
      ERGM        & $38.16(0.05)$ & $40.08(0.04)$ & $55.90(0.22)$\\
      VGAE        & $55.05(0.20)$ & $72.44(0.28)$ & $76.21(1.85)$\\
      NetGAN      & $49.34(0.22)$ & $63.62(0.32)$ & $66.96(5.77)$\\
      \midrule
      Improved HH & $\boldsymbol{74.15(0.13)}$ & $\boldsymbol{80.44(0.20)}$ & $\boldsymbol{80.84(0.17)}$\\
      \bottomrule
    \end{tabular}
  \end{adjustbox}
    
  \begin{adjustbox}{max width=\linewidth}
    \begin{tabular}{cccc}
      \toprule
      & \cora & \citeseer & \gene \\
      \midrule
      Origin. graph & $79.10(0.06)$ & $86.52(0.04)$ & $85.14(0.16)$\\
      \midrule
      CONF        & $35.61(0.12)$ & $51.99(0.06)$ & $64.37(0.69)$\\
      DC-SBM      & $56.72(0.10)$ & $60.63(0.05)$ & $69.54(0.18)$\\
      Chung-Lu    & $49.07(0.11)$ & $47.88(0.06)$ & $62.02(0.08)$\\
      ERGM        & $52.59(0.05)$ & $55.07(0.06)$ & $59.41(0.31)$\\
      VGAE        & $50.75(0.18)$ & $74.04(0.07)$ & $76.42(0.32)$\\
      NetGAN      & $44.10(0.19)$ & $67.97(0.04)$ & $68.30(0.15)$\\
      \midrule
      Improved HH & $\boldsymbol{78.07(0.10)}$ & $\boldsymbol{83.30(0.16)}$ & $\boldsymbol{80.89(0.12)}$\\
      \bottomrule
    \end{tabular}
  \end{adjustbox}
\end{table}

\textbf{Visualization of generated graphs.}
For qualitative evaluation, we visualize the original graph and two generated graphs: one by NetGAN, a recently proposed deep generative model for this setting; and the other by our model. The \cora\ graphs are given in Figure~\ref{cora_result} and the \citeseer\ graphs and \gene\ graphs are left in the supplement (Section~\ref{sec:remain}). Owing to the objective, NetGAN generates a graph considerably overlapping with the original one, with the delta edges scattering like noise. On the other hand, our model generates a graph that minimally overlaps with the original one. Hence, the drawing shows a quite different look. We highlight, however, that despite the dissimilar appearance, the many global and local properties remain close as can be seen from Table~\ref{metric_table_cora} (and~\ref{metric_table_citeseer} and~\ref{metric_table_gene} in the supplement).

\section{Conclusions}
Generating a new graph that barely overlaps with the input yet closely matches its properties has many practical uses, especially in proprietary settings. We show a random graph example that indicates that such graphs exist and we propose a method to generate them. The method is motivated by the Havel--Hakimi (HH) algorithm that preserves the degree sequence of the input, but improves its performance through introducing new node embeddings and using their link probabilities to guide the execution of HH. We empirically validate the method on three benchmark data sets, demonstrating appealing match of the graph properties, minimal overlap in edges, and similar performance in node classification.

\bibliography{doppelganger_icml}
\bibliographystyle{icml2021}

\clearpage
\appendix

\section{Proofs}\label{proofs}
\subsection{Proof of Theorem \ref{theorem 1}.}
\begin{proof}
Since the degree sequence of the origin graph is graphic, the Havel--Hakimi algorithm is guaranteed to generate a new graph. By construction, the new graph has the same degree sequence.

Let $V$ be the node set and $d(v)$ be the degree of a node $v$. The wedge count is defined as $\sum_{v \in V} \binom{d(v)}{2}$. The powerlaw exponent is $1+n(\sum_{v \in V}\log \frac{d(v)}{d_{\min}})^{-1}$, where $d_{\min}$ denotes the smallest degree. The entropy of the degree distribution is $\frac{1}{|V|}\sum_{v\in V}-\frac{d(v)}{|E|}\log\frac{d(v)}{|E|}$, where $E$ denotes the edge set. The Gini coefficient is $\frac{2\sum_{i=1}^{|V|i\widehat{d}_i}}{|V|\sum_{i=1}^{|V|}\widehat{d}_i}-\frac{|V|+1}{|V|}$, where $\widehat{d}$ is the list of sorted degrees. All these quantities are dependent solely on the degree sequence.
\end{proof}

\subsection{Proof of Theorem \ref{theorem 2}.}
\begin{proof}
Consider the smallest $k$ such that $d_1 \geq d_k \geq d_{k+1}+k-1$. Because $d_{k+1}+k-1 \geq k-1$, the first node must have a degree $\ge k-1$. Hence, the Havel--Hakimi algorithm will connect this node to the $2$nd, $3$rd, $\ldots$, $k$th nodes. Then, the remaining degrees become $d_2-1,d_3-1,....,d_k-1,d'_{k+1}, \ldots, d'_n)$, where $d'_{i} =d_i$ or $d_i-1$.

For the second node with $d_2-1$ remaining degrees, because $d_2-1 \ge k-2$, it must be connected to the $3$rd, $4$th, $\ldots$, $k$th nodes. Then, the remaining degrees  become $(d_3-2,....,d_k-2,d''_{k+1},.....,d''_n)$, where $d''_{i} =d'_i$ or $d'_i-1$.

After this process repeats  $k-1$ times, we can see that the first $k$ nodes have been connected to each other. As a result, the graph generated by the Havel--Hakimi algorithm must form a clique with at least $k$ nodes.
\end{proof}

\section{Code}
Our code is available at {\url{https://github.com/yizhidamiaomiao/DoppelgangerGraph}.}

\section{Baseline Models and Configurations}\label{sec:baseline}
To gauge the performance of the proposed method, we use NetGAN as well as a number of baselines introduced in~\citet{Bojchevski18NetGAN}. The configurations may differ from those described by the referenced work due to, e.g., software packages and implementation.

\textbf{Configuration model (CONF) \citep{Molloy1995}:} For a desired edge overlap, we sample the corresponding number of edges in the original graph and randomly rewire the rest of the edges. For uniformly random rewiring, there is a certain probability that all node degrees are preserved in the new graph. We repeat the generation until such a graph is produced. The larger the overlap, the higher the probability of successfully obtaining a new graph. We use the overlap $42.4\%$ appearing in~\citet{Bojchevski18NetGAN} for experimentation.

\textbf{Degree-correlated stochastic block model (DC-SBM) \citep{dcsbm2011}:} We use the Python package \texttt{graspy.models} and apply the default parameters.

\textbf{Exponential random graph model (ERGM) \citep{ERGM1981}:} We use the R function \texttt{ergm} from the \texttt{ergm} package~\citep{ergmpackage}. The only parameter is \texttt{edges}.

\textbf{Chung-Lu model~\citep{ChungLu2001}:} We implement the method based on the description of the paper. 

\textbf{Variational graph autoencoder (VGAE) \citep{kipf2016variational}:} We use the PyTorch implementation provided by the authors (\url{https://github.com/~tkipf/gae}). For each graph we train 2000 iterations and use the trained model to generate five edge probability matrices. We then use the method described in Section 3.3 of~\citet{Bojchevski18NetGAN} to generate five graphs.

\textbf{NetGAN \citep{Bojchevski18NetGAN}:} We use the Tensorflow implementation provided by the authors (\url{https://github.com/danielzuegner/netgan}) and apply the edge-overlap criterion for early stopping. We use hyperparameters provided therein.

\section{Graph Properties}\label{sec:property}
See Table~\ref{explanation of metrics} for a list of graph properties that we use to evaluate different methods. The last three are distributions and hence we compute the maximum mean discrepancy (MMD) between the distribution of the original graph and that of the generated graph.

\begin{table*}[ht!]
  \centering
  \caption{Graph properties.}
  \label{explanation of metrics}
  \renewcommand{\arraystretch}{1.3}
  \vskip.05in
  \begin{tabular}{p{2in}p{4.4in}}
    \toprule
    Property & Description\\
    \midrule
    Clustering coefficient & Number of closed triplets divided by number of all triplets.\\
    Characteristic path length &  The median of the means of the shortest path lengths connecting each node to all other nodes.\\
    Triangle count & Number of triangles in the graph.\\
    Square count & Number of squares in the graph.\\
    LCC & Size of the largest connected component.\\
    Power law exponent & Exponent of the power law distribution.\\
    Wedge count & Number of wedges (i.e., 2-stars; two-hop paths).\\
    Relative edge distribution entropy & Entropy of degree distribution.\\
    Gini coefficient & Common measure for inequality in a distribution.\\
    Local clustering coefficients & The coefficients form a distribution. We compute the MMD between the distribution of the original graph and that of the generated graph.\\
    Degree distribution & The degrees form a distribution. We compute the MMD between the distribution of the original graph and that of the generated graph.\\
    Local square clustering coefficients & The coefficients form a distribution. We compute the MMD between the distribution of the original graph and that of the generated graph.\\
    \bottomrule
  \end{tabular}
\end{table*}

\section{Our Model Configuration and Training}\label{sec:param}

\begin{table*}[ht!]
  \small
  \renewcommand{\arraystretch}{1.5}
  \centering
  \caption{Architectures and hyperparameters.}
  \label{tab:param}
  \vskip.05in
  \begin{tabular}{ll}
    \toprule
    \multirow{3}{*}{\cora}
    & GraphSAGE training: $C=1$, $T=20$, $E_0=5000$, $E_1=5000$, $K=2000$. \\
    & GAN generator: $FC(16,32) \to FC(32,64) \to FC(64,100) \to FC(100,128)$. \\
    & GAN discriminator: $FC(128,100) \to FC(100,64) \to FC(64,32) \to FC(32,1)$. \\
    \midrule
    \multirow{3}{*}{\citeseer}
    & GraphSAGE training: $C=2$, $T=20$, $E_0=2500$, $E_1=4000$, $K=1000$. \\
    & GAN generator: $FC(16,32) \to FC(32,64) \to FC(64,100) \to FC(100,128)$. \\
    & GAN discriminator: $FC(128,100) \to FC(100,64) \to FC(64,32) \to FC(32,1)$. \\
    \midrule
    \multirow{3}{*}{\gene}
    & GraphSAGE training: $C=1$, $T=20$, $E_0=2500$, $E_1=4000$, $K=1000$. \\
    & GAN generator: $FC(16,32) \to FC(32,64) \to FC(64,100) \to FC(100,128+2)$. \\
    & GAN discriminator: $FC(128+2,100) \to FC(100,64) \to FC(64,32) \to FC(32,1)$. \\
    \bottomrule
  \end{tabular}
\end{table*}

A high quality link predictor is key for the improved HH algorithm to select the right neighbors. A typical problem for link prediction is that negative links (unconnected node pairs) dominate and they scale quadratically with the number of nodes. A common remedy is to perform negative sampling during training, so that the training examples do not explode in size and remain relatively balanced. However, model evaluation is generally not properly conducted, because not every negative link is tested. As more negative links are evaluated, false positive rate tends to increase.

Thus, we enhance GraphSAGE training with a cycling approach ($C$ cycles of $T$ rounds). In each cycle, we start the first round by using the same number of positive and negative links. All negative links (hereby and subsequent) are randomly sampled. We run $E_0$ epochs for the first round. Afterward, we insert $K$ unused negative links into the training set and continue training with $E_1$ epochs, as the second round. In all subsequent rounds we similarly insert further $K$ unused negative links into the training set and train for $E_1$ epochs. With $T$ rounds we complete one cycle. Afterward, we warm start the next cycle with the trained model but a new training set, which contains the same number of positive and random negative links. Such a warm-start cycling avoids exploding the number of negative links used for training.

Details of the choice of $C$, $T$, $E_0$, $E_1$, and $K$ are given in Table~\ref{tab:param}. In this table, we also summarize the architectures and hyperparameters for GAN.

\section{Link Prediction and GAN Training Results}

\begin{figure*}[ht]
  \centering
  \includegraphics[width=0.33\linewidth]{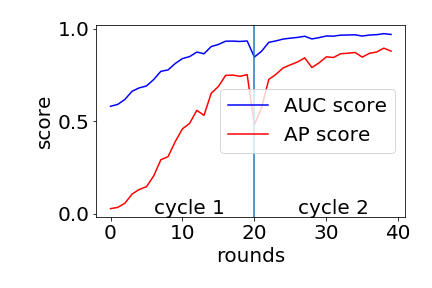}
  \includegraphics[width=0.33\linewidth]{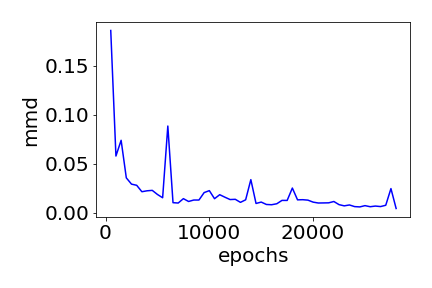}
  \includegraphics[width=0.32\linewidth]{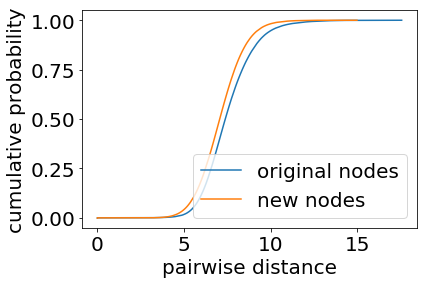} \\
  \includegraphics[width=0.33\linewidth]{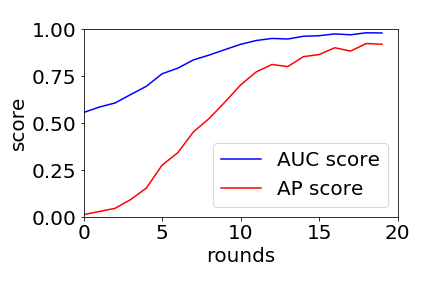}
  \includegraphics[width=0.33\linewidth]{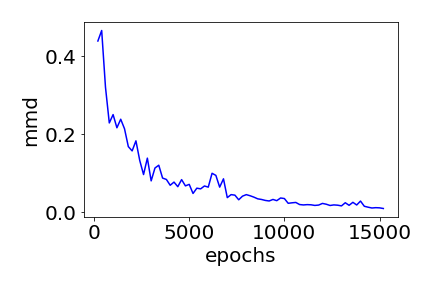}
  \includegraphics[width=0.32\linewidth]{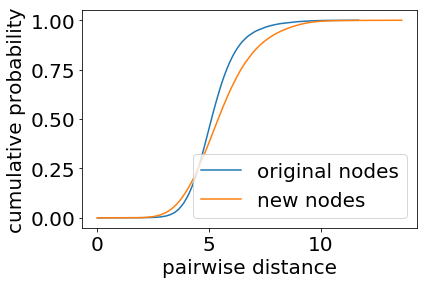} \\
  \includegraphics[width=0.33\linewidth]{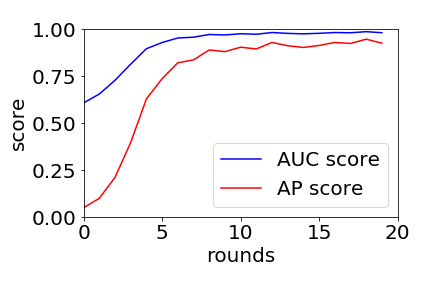}
  \includegraphics[width=0.33\linewidth]{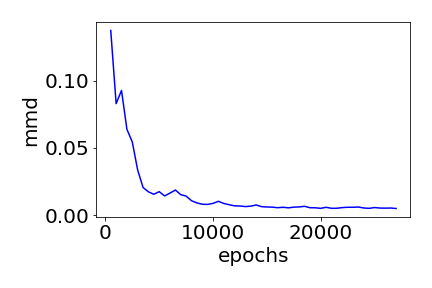}
  \includegraphics[width=0.32\linewidth]{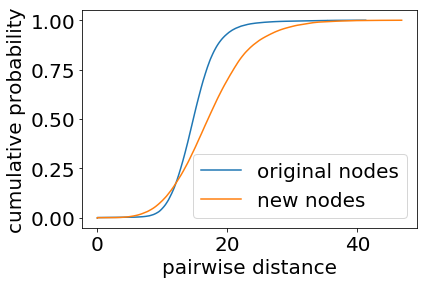} \\
  \caption{Top: \cora; Middle: \citeseer. Bottom: \gene. Left: GraphSAGE training progress (AP, AUC); middle: GAN training progress (MMD); right: GAN result (pairwise distance distribution).}
  \label{GraphSAGE_linkpred_GAN_cora}
\end{figure*}

Here, we inspect the training of the link predictor. The result of the cycling approach described in the preceding section is illustrated on the left panel of Figure~\ref{GraphSAGE_linkpred_GAN_cora}. The metrics are AP and AUC scores evaluated on all node pairs. For \cora, we use two cycles. One sees that the second cycle improves over the score at the end of the first cycle, although the warm start begins at a lower position. For \citeseer\ and \gene, we use only one cycle, because the AUC score is already close to full at the end of the cycle. For both data sets, the purpose of incrementally inserting more negative edges in each round is fulfilled: scores progressively increase. Had we not inserted additional negative edges over time, the curves would plateau after the first round. Overall, the convergence history indicates that the enhanced GraphSAGE training strategy is quite effective.

We next inspect the training of GAN. GAN results are known to be notoriously hard to robustly evaluate. Here, we show the MMD; see the middle panel of Figure~\ref{GraphSAGE_linkpred_GAN_cora}. It decreases generally well, with small bumps throughout. The challenge is that it is hard to tell from the value whether training is sufficient. Therefore, we also investigate the empirical cdf of the pairwise data distance. As shown in the right panel of Figure~\ref{GraphSAGE_linkpred_GAN_cora}, the cdf for the training data and that for the generated data are visually close. Such a result indicates that GAN has been trained reasonably well.

\section{Node Classification Details}\label{sec:node.classi}
We perform node classification with two models: GCN\footnote{\url{https://github.com/tkipf/pygcn}} and GraphSAGE\footnote{\url{https://github.com/williamleif/graphsage-simple/}}. The codes are downloaded from links given in the footnotes and we use the default hyperparameters without tuning.

\cora\ and \citeseer\ come with node features while \gene\ not. For \gene, following convention we use the identity matrix as the feature matrix. For all compared methods, the node set is fixed and hence the node features are straightforwardly used when running GCN and GraphSAGE; whereas for our method, because nodes are new, we use their embedding generated by GAN as the input node features. To obtain node labels, we augment GAN to learn and generate them. Specifically, we concatenate the node embeddings with a one-hot vector of the class label as input/output of GAN. It is senseless to use a fixed train/test split, which is applicable to only the original node set. Hence, we conduct ten-fold cross validation to obtain classification performance.

Note that each generative model can generate an arbitrary number of graphs. In practice, one selects one or a small number of them to use. In Table~\ref{node_classification}, we select the best one based on cross validation performance after 20 random trials.

\section{Remaining Experiment Results}\label{sec:remain}
\begin{figure*}[ht!]
  \centering
  \includegraphics[width=.32\linewidth]{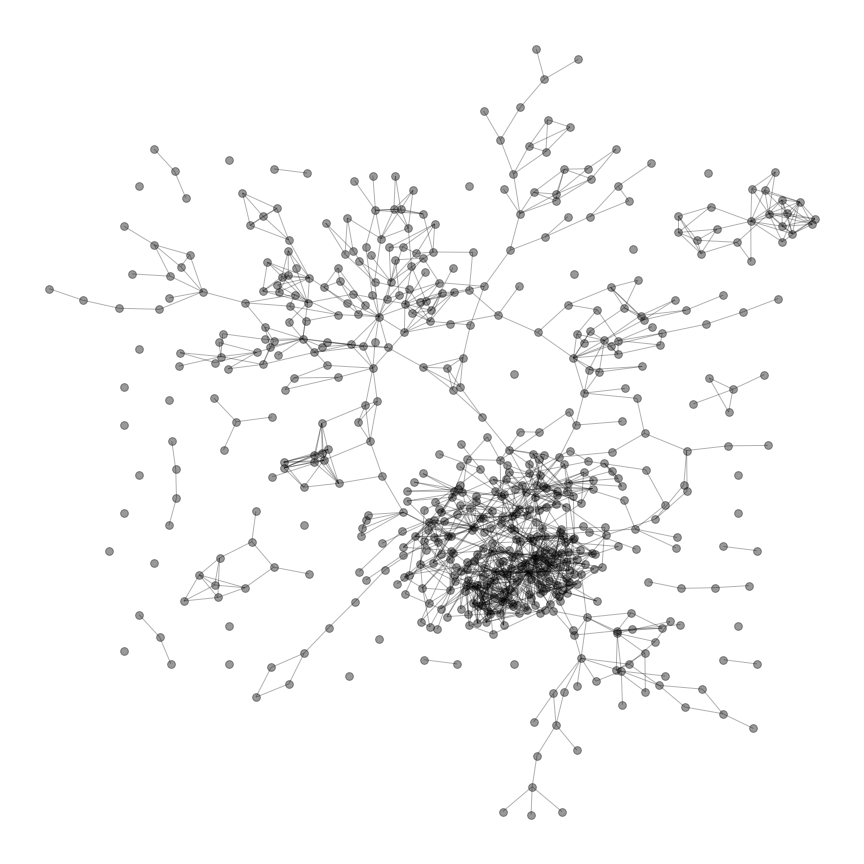}
  \includegraphics[width=.32\linewidth]{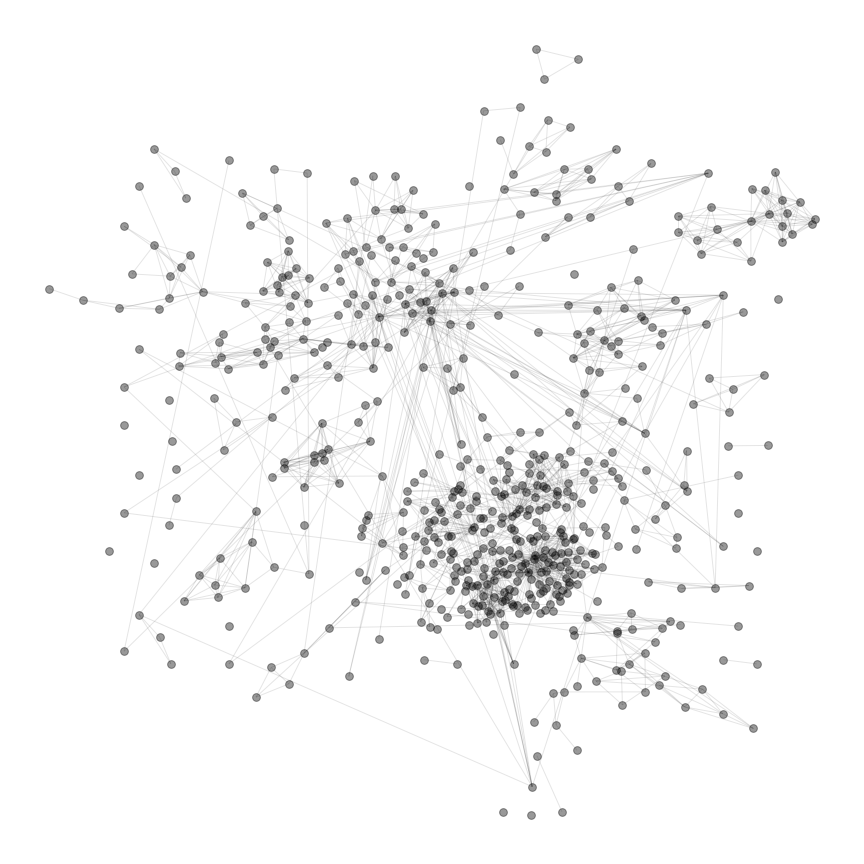}
  \includegraphics[width=.32\linewidth]{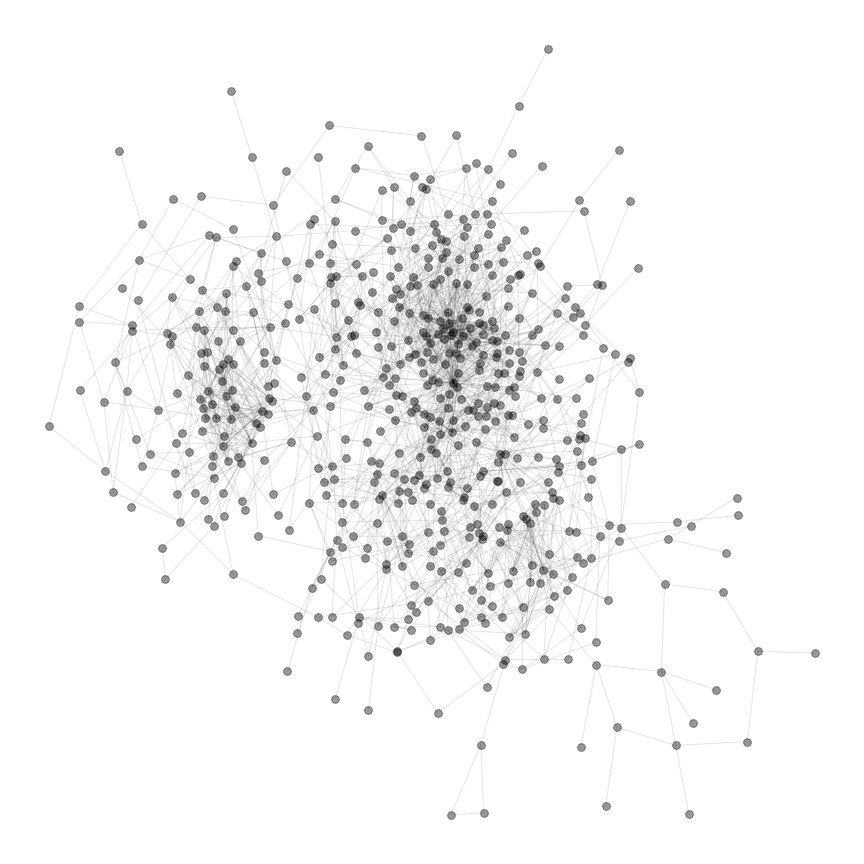}
  \caption{\citeseer. Left: original graph; middle: generated by NetGAN; right: generated by improved HH. Drawn are the subgraphs induced by the first 600 nodes ordered by degrees.}
  \label{citeseer_result}
\end{figure*}

\begin{figure*}[ht!]
  \centering
  \includegraphics[width=.32\linewidth]{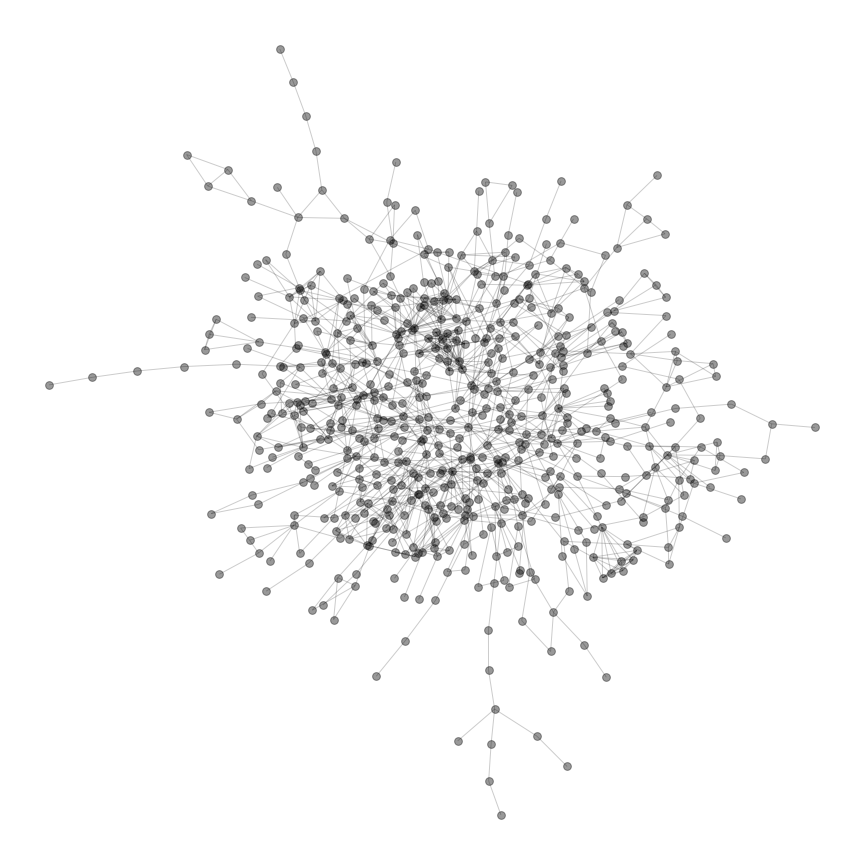}
  \includegraphics[width=.32\linewidth]{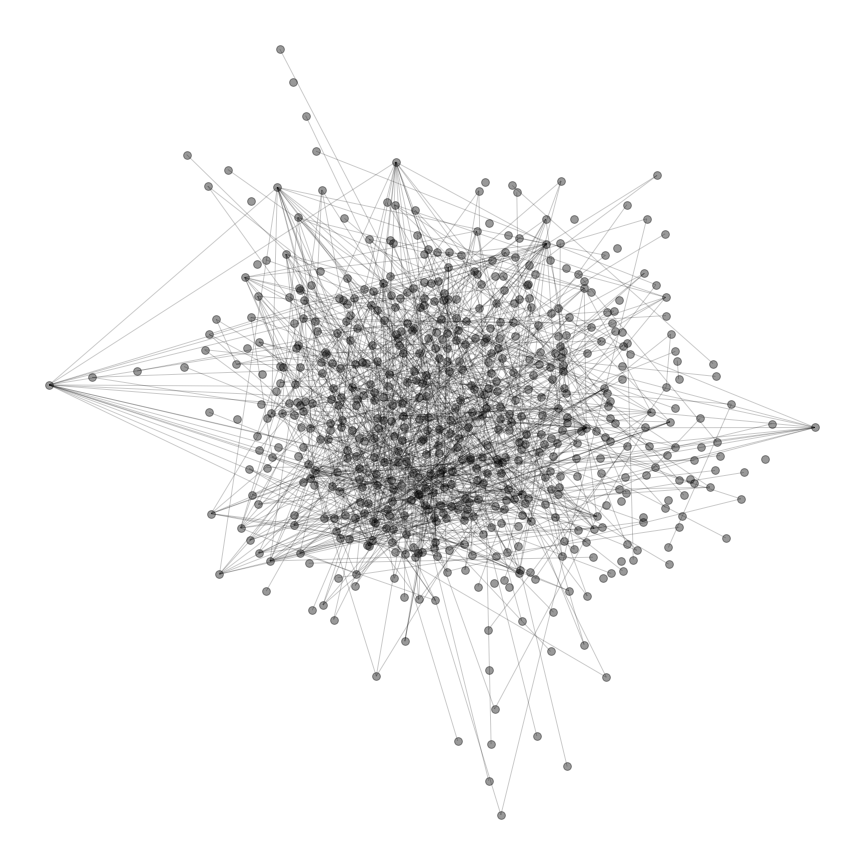}
  \includegraphics[width=.32\linewidth]{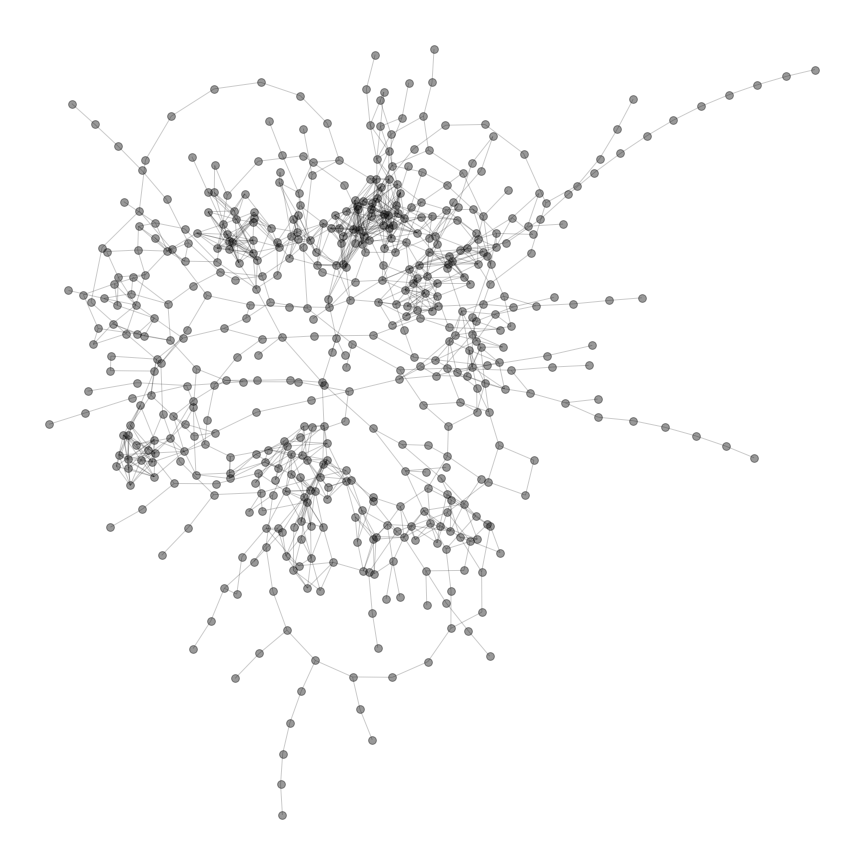}
  \caption{\gene. Left: original graph; middle: generated by NetGAN; right: generated by improved HH. Drawn are the subgraphs induced by the first 600 nodes ordered by degrees.}
  \label{gene_result}
\end{figure*}

In the main text, we show the quality comparison and visualization results for only \cora. Here, results for \citeseer\ are given in Table~\ref{metric_table_citeseer} and Figures~\ref{citeseer_result}, whereas results for \gene\ are given in Table~\ref{metric_table_gene} and Figure~\ref{gene_result}.

\begin{table*}[ht!]
  \small
  \centering
  \caption{Properties of \citeseer\ and the graphs generated by different models (averaged over five random repetitions). EO means edge overlap and numbers in the bracket are standard deviations.}
  \label{metric_table_citeseer}
  \vskip.05in
  \begin{adjustbox}{max width=\textwidth}
    \begin{tabular}{c@{\hspace{.5em}}ccccccc}
      \toprule
      & EO & Cluster. coeff. & Charcs. path length & Triangle count& Square count & LCC & Powerlaw exponent  \\
      & & $\times10^{-2}$ & $\times10^{+0}$ & $\times10^{+3}$ & $\times10^{+2}$ & $\times10^{+3}$ & $\times10^{+0}$\\
      \midrule
      \citeseer   &          & $1.30$ & $9.33$ & $1.08$ & $2.49$ & $2.12$ & $2.07$\\
      \midrule
      CONF        & $42.4\%$ & $0.160(0.019)$ & $5.39(0.01)$ & $0.133(0.016)$ & $0.02(0.02)$ & $\boldsymbol{2.09(0.01)}$ & $\boldsymbol{2.07(0.00)}$\\
      DC-SBM      & $5.35\%$ & $0.755(0.065)$ & $5.58(0.09)$ & $0.500(0.025)$ & $0.794(0.134)$ & $1.75(0.02)$ & $1.95(0.01)$\\
      Chung-Lu    & $0.86\%$ & $0.0786(0.0088)$ & $4.73(0.05)$ & $0.076(0.005)$& $0.006(0.008)$ & $1.80(0.01)$ & $1.94(0.02)$ \\
      ERGM        & $8.30\%$ & $0.18(0.036)$ & $6.15(0.04)$ & $0.00940(0.00196)$ & $0.00(0.00)$ & $2.04(0.00)$ & $1.87(0.00)$\\
      VGAE        & $45.8\%$ & $2.92(0.14)$ & $\boldsymbol{8.66(0.87)}$ & $6.94(0.14)$  & $133(11)$ & $0.90(0.129)$ & $1.86(0.01)$ \\
      NetGAN      & $55.9\%$ & $2.42(0.70)$ & $6.75(0.63)$ & $4.53(0.18)$ & $45.8(5.0)$ & $1.36(0.04)$ & $1.81(0.01)$ \\
      \midrule
      Improved HH & $\boldsymbol{0.11\%}$ & $\boldsymbol{0.839(0.057)}$ & $7.64(0.29)$ & $\boldsymbol{0.700(0.048)}$ & $\boldsymbol{2.10(0.72)}$ & $1.67(0.02)$ & $\boldsymbol{2.07(0.00)}$ \\
      \bottomrule
      \toprule
      & EO & Wedge count & Rel. edge distr. entr. & Gini coefficient & Local cluster. & Degree distr. &  Local sq. cluster. \\
      & & $\times10^{+4}$ & $\times10^{-1}$ & $\times10^{-1}$ & $\times10^{-2}$ & $\times10^{-2}$ & $\times10^{-3}$\\
      \midrule
      \citeseer   &          & $2.60$ & $9.54$ & $4.28$ & $0$ & $0$ & $0$\\
      \midrule
      CONF        & $42.4\%$ & $\boldsymbol{2.60(0.00)}$ & $\boldsymbol{9.54(0.00)}$ & $\boldsymbol{4.28(0.00)}$& $3.20(0.07)$ & $\boldsymbol{0}$ & $7.26(0.26)$  \\
      DC-SBM      & $5.35\%$ & $2.73(0.04)$ & $9.34(0.01)$ & $5.16(0.04)$ & $\boldsymbol{2.82(0.10)}$ & $3.15(0.12)$ & $6.72(0.21)$ \\
      Chung-Lu    & $0.86\%$ & $3.00(0.12)$ & $9.33(0.01)$ & $5.18(0.04)$ & $3.65(0.03)$ & $3.15(0.19)$ & $7.54(0.07)$\\
      ERGM        & $8.30\%$ & $1.31(0.02)$ & $9.79(0.00)$ & $2.98(0.02)$& $3.83(0.02)$ & $5.90(0.25)$ & $7.66(0.03)$  \\
      VGAE        & $45.8\%$ & $5.38(0.08)$ & $8.60(0.00)$ & $7.15(0.01)$ & $3.60(0.06)$ & $19.41(0.62)$ & $33.59(2.24)$\\
      NetGAN      & $55.9\%$ & $4.21(0.30)$ & $8.96(0.03)$ & $6.40(0.10)$ & $5.10(0.28)$ & $16.26(0.58)$ & $15.47(2.15)$\\
      \midrule
      Improved HH & $\boldsymbol{0.11\%}$ & $\boldsymbol{2.60(0.00)}$ & $\boldsymbol{9.54(0.00)}$ & $\boldsymbol{4.28(0.00)}$ & $3.20(0.04)$ & $\boldsymbol{0}$ & $\boldsymbol{4.71(0.19)}$\\
      \bottomrule
    \end{tabular}
  \end{adjustbox}

  \vskip30pt
  \includegraphics[width=\linewidth]{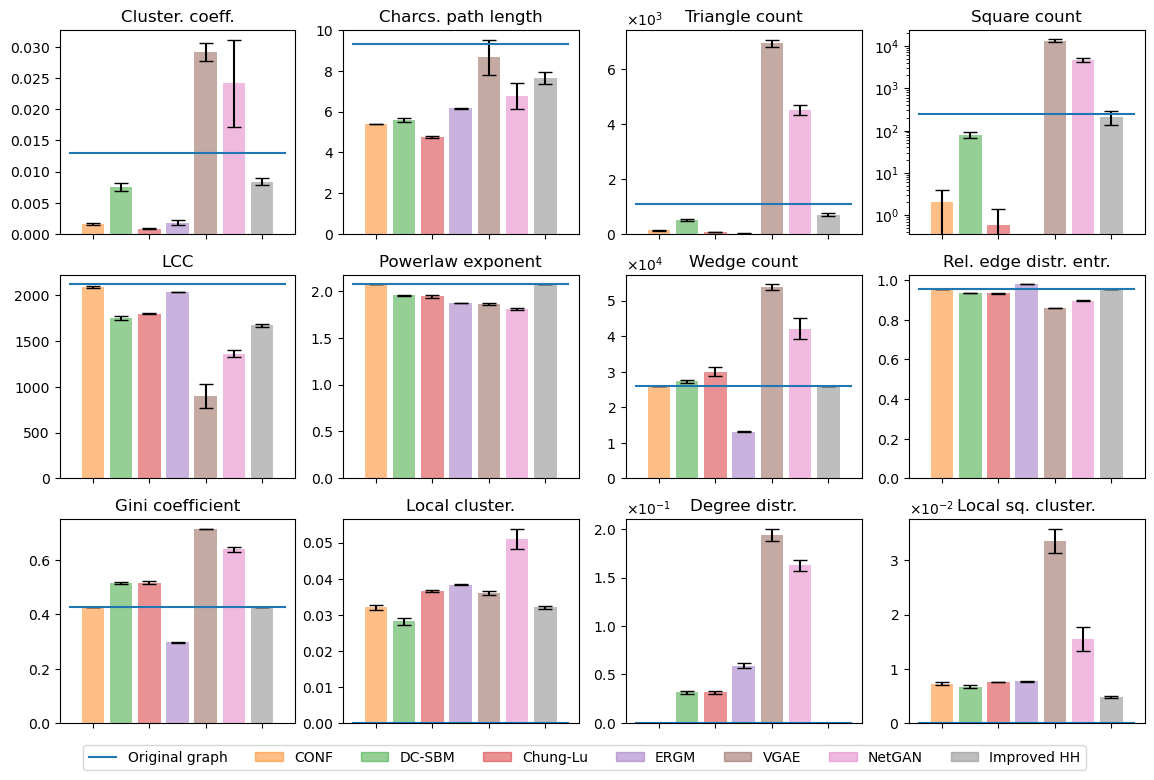}
\end{table*}

\begin{table*}[ht!]
  \small
  \centering
  \caption{Properties of \gene\ and the graphs generated by different models (averaged over five random repetitions). EO means edge overlap and numbers in the bracket are standard deviations.}
  \label{metric_table_gene}
  \vskip.05in
  \begin{adjustbox}{max width=\textwidth}
    \begin{tabular}{c@{\hspace{.5em}}ccccccc}
      \toprule
      & EO & Cluster. coeff. & Charcs. path length & Triangle count& Square count & LCC & Powerlaw exponent  \\
      & & $\times10^{-2}$ & $\times10^{+0}$ & $\times10^{+2}$ & $\times10^{+2}$ & $\times10^{+2}$ & $\times10^{+0}$\\
      \midrule
      \gene       &          & $10.41$ & $7.01$ & $8.09$ & $9.68$ & $8.14$ & $2.05$\\
      \midrule
      CONF        & $42.7\%$ & $1.21(0.06)$ & $5.01(0.05)$ & $0.94(0.04)$ & $0.11(0.04)$ & $\boldsymbol{7.98(0.06)}$ & $\boldsymbol{2.05(0.00)}$\\
      DC-SBM      & $11.8\%$ & $4.36(0.38)$ & $5.01(0.06)$ & $3.46(0.36)$ & $3.09(0.78)$ & $6.81(0.14)$ & $1.92(0.03)$\\
      Chung-Lu    & $1.52\%$ & $0.382(0.058)$ & $4.33(0.03)$ & $0.42(0.07)$& $0.002(0.004)$ & $6.90(0.06)$ & $1.91(0.01)$ \\
      ERGM        & $0.60\%$ & $0.374(0.146)$ & $5.24(0.06)$ & $0.08(0.03)$ & $0.00(0.00)$ & $7.92(0.02)$ & $1.84(0.02)$\\
      VGAE        & $64.9\%$ & $13.8(0.05)$ & $\boldsymbol{6.99(0.19)}$ & $17.2(0.40)$  & $19.81(1.07)$ & $5.44(0.22)$ & $1.86(0.01)$ \\
      NetGAN      & $51.6\%$ & $7.54(1.18)$ & $4.50(0.31)$ & $22.53(1.99)$ & $28.78(5.95)$ & $4.60(0.28)$ & $1.72(0.03)$ \\
      \midrule
      Improved HH & $\boldsymbol{0.43\%}$ & $\boldsymbol{8.41(0.25)}$ & $10.72(0.87)$ & $\boldsymbol{6.54(0.19)}$ & $\boldsymbol{4.18(0.73)}$ & $6.32(0.25)$ & $\boldsymbol{2.05(0.00)}$ \\
      \bottomrule
      \toprule
      & EO & Wedge count & Rel. edge distr. entr. & Gini coefficient & Local cluster. & Degree distr. &  Local sq. cluster. \\
      & & $\times10^{+3}$ & $\times10^{-1}$ & $\times10^{-1}$ & $\times10^{-2}$ & $\times10^{-2}$ & $\times10^{-3}$\\
      \midrule
      \gene       &          & $7.79$ & $9.53$ & $4.26$ & $0$ & $0$ & $0$\\
      \midrule
      CONF        & $42.7\%$ & $\boldsymbol{7.79(0.00)}$ & $\boldsymbol{9.53(0.00)}$ & $\boldsymbol{4.26(0.00)}$& $5.39(0.19)$ & $\boldsymbol{0}$ & $14.24(0.28)$ \\
      DC-SBM      & $11.8\%$ & $8.20(0.33)$ & $9.35(0.03)$ & $4.95(0.01)$ & $4.10(0.06)$ & $2.71(0.55)$ & $10.72(0.20)$ \\
      Chung-Lu    & $1.52\%$ & $9.49(0.39)$ & $9.30(0.01)$ & $5.09(0.04)$ & $6.12(0.14)$ & $2.49(0.25)$ & $14.54(0.25)$\\
      ERGM        & $0.60\%$ & $5.51(0.18)$ & $9.78(0.01)$ & $2.86(0.04)$& $6.41(0.06)$ & $7.55(0.83)$ & $14.79(0.12)$  \\
      VGAE        & $64.9\%$ & $10.48(1.58)$ & $9.09(0.01)$ & $5.79(0.05)$ & $2.27(0.13)$ & $8.34(0.21)$ & $14.47(1.68)$\\
      NetGAN      & $51.6\%$ & $15.40(1.20)$ & $8.62(0.10)$ & $6.91(0.21)$ & $\boldsymbol{1.62(0.27)}$ & $23.5(3.21)$ & $6.89(1.73)$\\
      \midrule
      Improved HH & $\boldsymbol{0.43\%}$ & $\boldsymbol{7.79(0.00)}$ & $\boldsymbol{9.53(0.00)}$ & $\boldsymbol{4.26(0.00)}$ & $3.45(0.15)$ & $\boldsymbol{0}$ & $\boldsymbol{2.02(0.54)}$\\
      \bottomrule
    \end{tabular}
  \end{adjustbox}

  \vskip30pt
  \includegraphics[width=\linewidth]{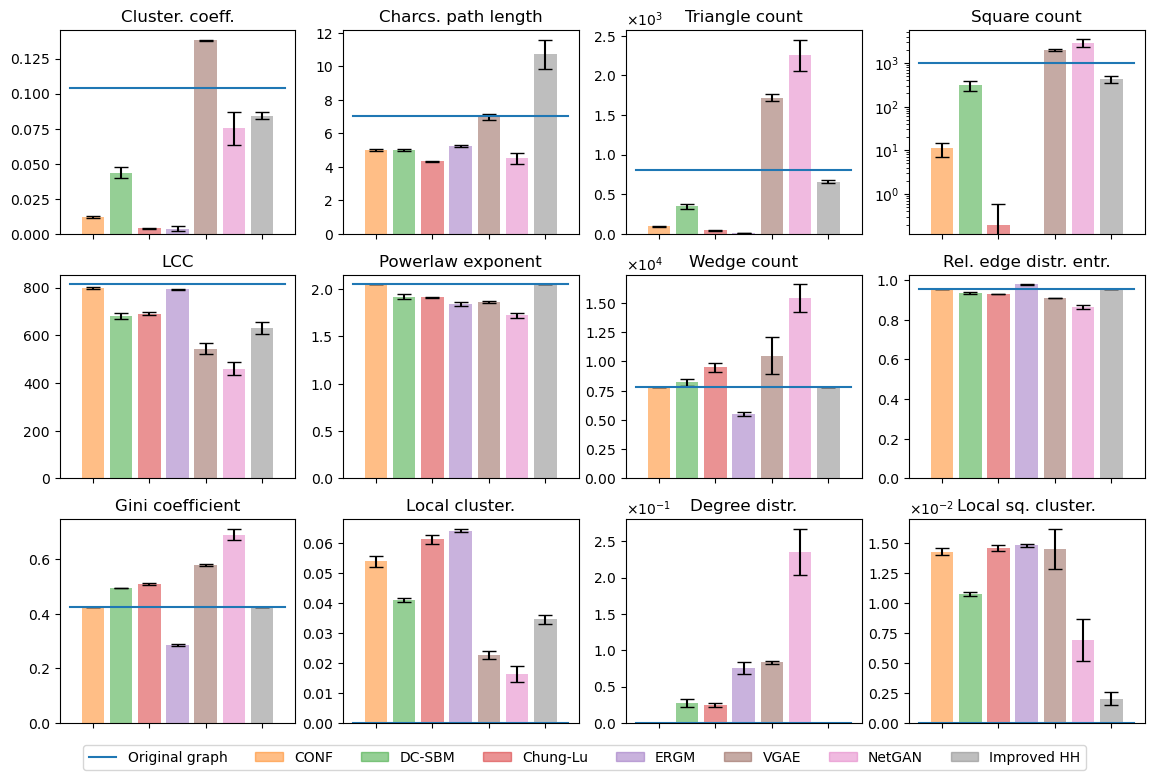}
\end{table*}

%



\end{document}